\documentclass[10pt,twocolumn,letterpaper]{article}

\usepackage{iccv}              % To produce the CAMERA-READY version
\definecolor{iccvblue}{rgb}{0.21,0.49,0.74}
\usepackage[pagebackref,breaklinks,colorlinks,allcolors=iccvblue]{hyperref}

\title{Local Information Matters: Inference Acceleration For  Grounded Conversation Generation Models Through Adaptive Local-Aware Token Pruning
}

\author{
Bizhe Bai$^{1,2}$ \quad Jianjian Cao$^{1}$   \quad Yadan Luo $^{3}$ \quad Tao Chen$^{1,2}$\thanks{Corresponding Author: \textit{eetchen@fudan.edu.cn}} \\ 
$^{1}$Fudan University \quad $^{2}$Shanghai Innovation Institute \\ 
\quad $^{3}$The University of Queensland \\
{\tt\small \quad eetchen@fudan.edu.cn}
}
\begin{document}
\maketitle
\begin{abstract}
Grounded Conversation Generation (GCG) is an emerging vision-language task that requires models to generate  natural language responses seamlessly intertwined with corresponding object segmentation masks. Recent models, such as GLaMM and OMG-LLaVA, achieve pixel-level grounding but incur significant computational costs due to processing a large number of visual tokens. Existing token pruning methods, like FastV and PyramidDrop, fail to preserve the local visual features critical for accurate grounding, leading to substantial performance drops in GCG tasks. To address this, we propose \textbf{Adaptive Local-Aware Token Pruning (ALTP)}, a simple yet effective framework that accelerates GCG models by prioritizing local object information. ALTP introduces two key components: (1) \textbf{Detail Density Capture (DDC)}, which uses superpixel segmentation to retain tokens in object-centric regions, preserving fine-grained details, and (2) \textbf{Dynamic Density Formation (DDF)}, which dynamically allocates tokens based on information density, ensuring higher retention in semantically rich areas. Extensive experiments on the GranDf dataset demonstrate that ALTP significantly outperforms existing token pruning methods, such as FastV and PyramidDrop, on both GLaMM and OMG-LLaVA models. Notably, when applied to GLaMM, ALTP achieves a 90\% reduction in visual tokens with a \textbf{4.9\%} improvement in AP50 and a \textbf{5.0\%} improvement in Recall compared to PyramidDrop. Similarly, on OMG-LLaVA, ALTP improves AP by \textbf{2.1\%} and mIOU by \textbf{3.0\%} at a 90\% token reduction compared with PDrop.
\end{abstract}    
\section{Introduction}
\label{sec:intro}

Grounded Conversation Generation (GCG) is an emerging vision-language task in which a model must produce textual responses accompanied by segmentation masks for specific regions of an image~\cite{rasheed2024glammpixelgroundinglarge}. For example, a GCG model might generate the sentence ``The bed is positioned beside the wall. A curtain is hanging from the wall,'' while also providing segmentation masks that precisely outline the bed, wall, and curtain. Compared to traditional vision-language tasks such as image captioning or visual question answering (VQA)~\cite{liu2023visualinstructiontuningllava, chen2024internvlscalingvisionfoundation, chen2025expandingperformanceboundariesopensourceinternvl2, wu2024deepseekvl2mixtureofexpertsvisionlanguagemodels, qwen2025qwen25technicalreport}, which only output text, or referring segmentation tasks~\cite{lai2024lisareasoningsegmentationlarge, chng2024maskgroundingreferringimage, xia2024gsvageneralizedsegmentationmultimodal}, which only predict segmentation masks for a given query phrase, GCG explicitly requires the model to produce holistic descriptions that interleave object mentions with their corresponding masks. This capability is vital for applications requiring precise visual reasoning, such as interactive agents that can identify objects in real-time or systems that generate detailed instructions for image editing.

Recently, Rasheed \emph{et al.} introduced GLaMM~\cite{rasheed2024glammpixelgroundinglarge}, the first large-scale model for GCG, demonstrating the feasibility of pixel-level grounding within multimodal conversations. OMG-LLaVA~\cite{zhang2024omgllavabridgingimagelevelobjectlevel} followed with a more compact architecture, achieving higher accuracy without relying on additional segmentation models like SAM~\cite{kirillov2023segment}. Both approaches generate complete image descriptions intertwined with segmentation masks (dense grounding) by processing a substantial number of visual tokens. For example, GLaMM employs 576 visual tokens per image, while OMG-LLaVA uses 256. Although these models achieve fine-grained accuracy, they incur substantial computational overhead. Since both are built upon the LLaVA framework~\cite{liu2023visualinstructiontuningllava}, one might consider existing token-pruning methods such as FastV~\cite{chen2024imageworth12tokensfastv} or PyramidDrop~\cite{xing2025pyramiddropacceleratinglargevisionlanguage} to reduce this cost.
\begin{figure*}[htbp] 
    \centering 
        \includegraphics[width=1\linewidth]{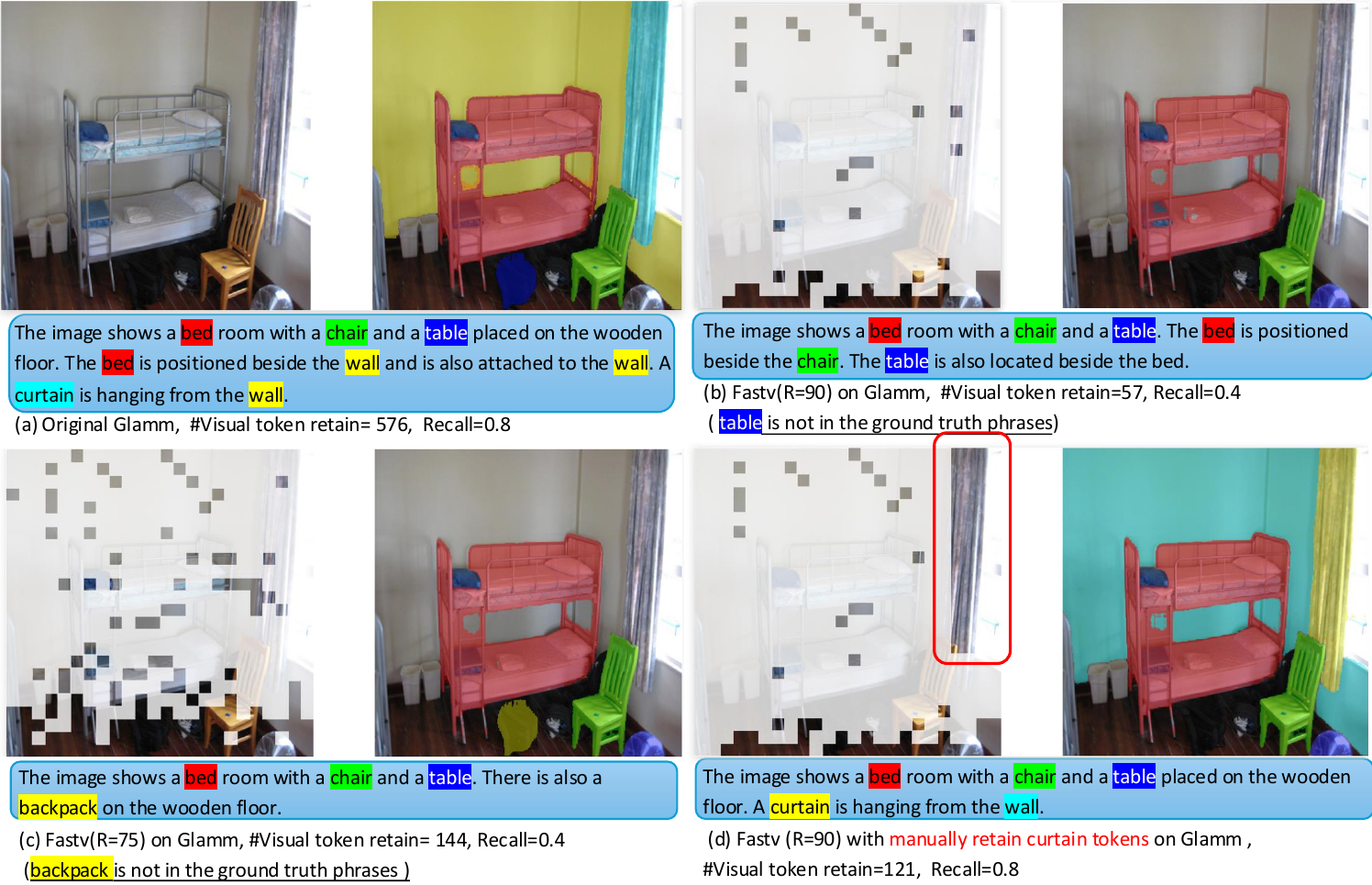} 
    \vspace{-5mm}
        \caption{(a): Result of original Glamm \cite{rasheed2024glammpixelgroundinglarge}. (b): Result of  Glamm with Fastv \cite{chen2024imageworth12tokensfastv} method pruning 90\% of visual token starting  from the second layer. (c) : Result from  Glamm with Fastv method pruning 75\% of visiual tokens starting from the second layer.  (d): Results from GlaMM using FastV \cite{chen2024imageworth12tokensfastv} with 90\% visual token pruning and retaining additional visual tokens retained at the curtain location, starting from the second layer. Comparing (c) and (d), we could conclude that \textbf{``Local Information Matters"} : preserving visual tokens corresponding to object locations provides richer object information to the vision-language model. }  
        \label{fig:glamm_pre_resul} 
    \vspace{-7mm}
\end{figure*}
However, when we applied FastV directly to GLaMM and OMG-LLaVA, the performance on GCG tasks dropped significantly. For instance, Figure~\ref{fig:glamm_pre_resul}(a)--(c) shows that pruning 75\% of GLaMM's visual tokens with FastV cut its recall by half, from $0.8$ to $0.4$. By contrast, using the same pruning ratio in text-focused multimodal tasks (e.g., MMMU~\cite{yue2024mmmumassivemultidisciplinemultimodal} or A-OKVQA~\cite{marino2019okvqavisualquestionanswering}) reduced performance by only about 2\%. We hypothesize that GCG tasks demand more localized visual features than traditional VQA or captioning tasks. Inspired by Pan \emph{et al.}~\cite{pan2023slidetransformerhierarchicalvisiontransformer}, who showed that global attention struggles to capture local details in vision transformers, we further tested whether reintroducing tokens corresponding to small, salient objects (e.g., a curtain) could restore GCG performance. Indeed, selectively retaining these local tokens not only recovered lost accuracy but sometimes surpassed the performance of less aggressive pruning schemes (Figure~\ref{fig:glamm_pre_resul}(d)).

\begin{figure*}[htbp] 
    \centering 
        \includegraphics[width=1\linewidth]{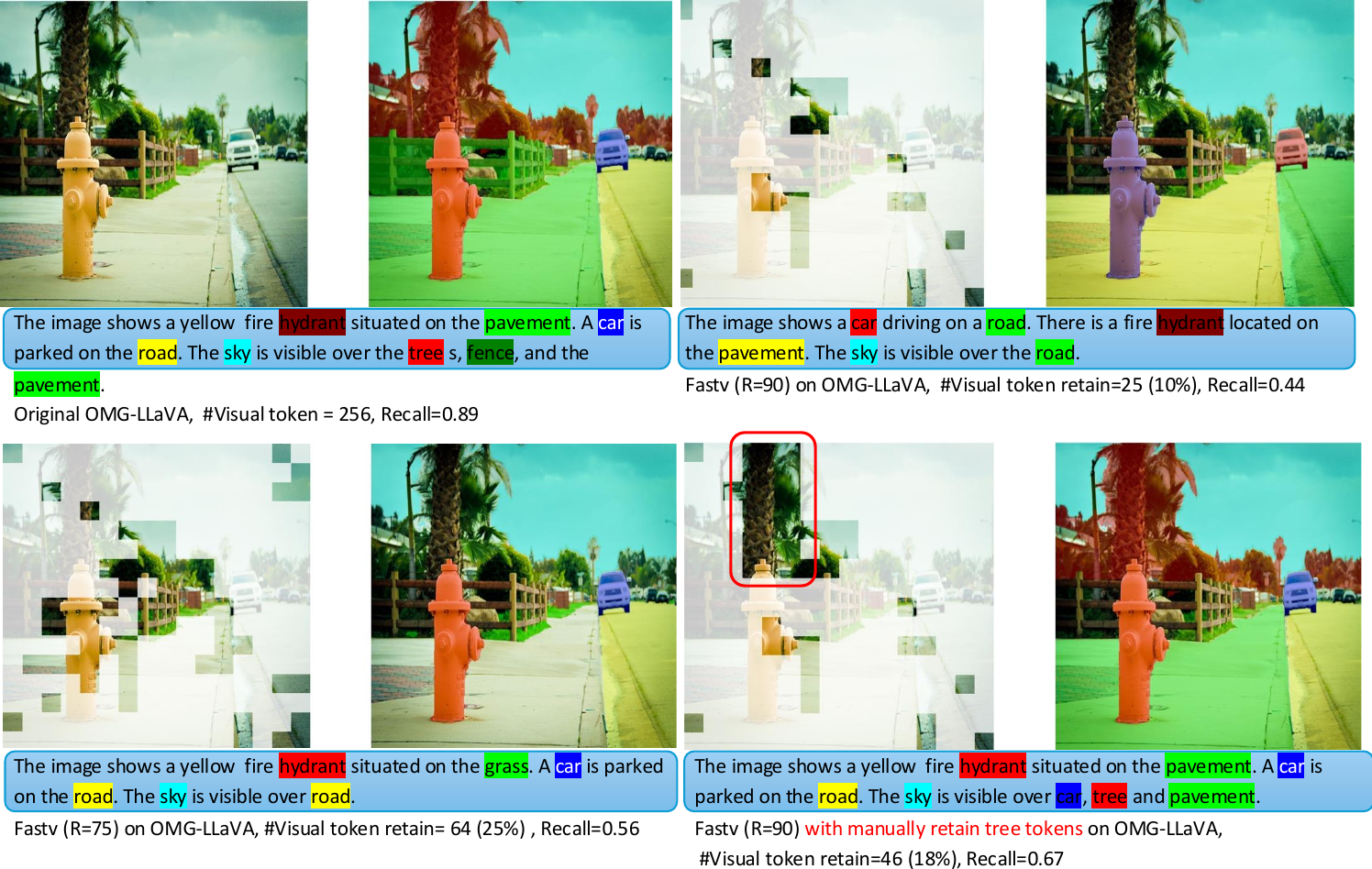} 
    \vspace{-6mm}
        \caption{(a): Result of original OMG-Llava \cite{zhang2024omgllavabridgingimagelevelobjectlevel}. (b): Result of  OMG-Llava with Fastv \cite{chen2024imageworth12tokensfastv} method pruning 90\% of visual token starting  from the second layer. (c) Result of pruning 75\% of visual tokens starting from the second layer. (d) Pruning 90\% visual token pruning and retaining additional visual tokens retained at the tree location. Comparing (c) and (d), we can conclude the same conclusion  as Figure \ref{fig:glamm_pre_resul}: “Local Information Matters” . }  
        \label{fig:glamm_pre_resul} 
    \vspace{-6mm}
\end{figure*}
Drawing on these observations, we propose a novel token-pruning approach tailored for GCG models. Rather than relying exclusively on global cross-attention to identify redundant tokens, our framework explicitly preserves tokens that encode local, object-centric information. Concretely, we make three key contributions:

\begin{itemize}[leftmargin=*]
    \item We highlight why standard token-pruning methods struggle in GCG tasks, emphasizing the importance of local visual features in grounded conversation.
    \item We propose a simple yet effective framework that combines superpixel segmentation with an adaptive token-allocation strategy, ensuring critical object-level details are retained.
    \item Through extensive experiments on the GranDf dataset, we demonstrate that our method significantly outperforms existing token-pruning approaches (FastV, PyramidDrop) on both GLaMM and OMG-LLaVA, reducing token counts by up to 90\% while improving segmentation and recall metrics.
\end{itemize}

\section{Related Work}
\subsection{Multimodal Large Language Models}
The field of Multimodal Large Language Models (MLLMs) has witnessed explosive growth in recent years, driven by advancements in deep learning and the increasing availability of multimodal datasets. 
Early works like Flamingo~\cite{alayrac2022flamingo} and BLIP-2~\cite{li2023blip} leveraged pre-trained vision encoders and LLMs, connected via lightweight adapters, to achieve impressive performance on tasks like image captioning and visual question answering.
In contrast, LLaVA~\cite{liu2023llava} and LLaVA-1.5~\cite{liu2024improved} employ the MLP layer to directly map encoded visual features into the input embedding space of the LLM, significantly simplifying both the model architecture and training pipeline. 
Despite these advancements, most existing MLLMs are primarily designed to generate textual outputs from multimodal inputs but lack the capability to predict object masks in images, limiting their effectiveness in fine-grained visual understanding tasks. 
%Additionally, their large parameter sizes and high computational requirements create significant deployment challenges. 
%To address these limitations, this work focuses on accelerating fine-grained multimodal models, aiming to enhance efficiency without compromising performance.

\subsection{Grounded Conversation Generation}
Recent research in GCG task has focused on enabling region-specific dialogue by integrating spatial and visual information with language models. Several notable works~\cite{peng2023kosmos,chen2023shikra,zhang2023gpt4roi,wang2023visionllm} have explored methods to facilitate fine-grained, region-level understanding in multimodal conversations. These approaches typically incorporate location bins, bounding boxes, or spatial coordinates alongside image data to enhance the model's ability to interpret and reason about specific regions within an image. For instance, methods like Shikra~\cite{chen2023shikra} and Kosmos-2~\cite{peng2023kosmos} rely on large language models (LLMs) to process region-specific inputs, such as bounding boxes, and generate corresponding textual descriptions.
GPT4RoI~\cite{zhang2023gpt4roi} advances this paradigm by incorporating spatial boxes and region-of-interest (RoI)-aligned features as inputs, training on region-text pairs to improve alignment between visual regions and textual outputs. Similarly, BuboGPT~\cite{zhao2023bubogpt} leverages an off-the-shelf grounding model~\cite{liu2024grounding} to align visual groundings with language responses, enabling more precise region-aware dialogue generation. In contrast, LISA~\cite{lai2024lisa} utilizes embeddings from a vision-language model combined with the Segment Anything Model (SAM)~\cite{kirillov2023segment} decoder to generate segmentation masks as outputs.
More recently, GLaMM~\cite{rasheed2024glammpixelgroundinglarge} has demonstrated the feasibility of pixel-level grounding in visual-language conversations, enabling fine-grained understanding of images.
Building on this, OMG-LLaVA~\cite{zhang2024omgllavabridgingimagelevelobjectlevel} proposed a more streamlined framework, achieving superior accuracy without relying on external segmentation models like SAM~\cite{kirillov2023segment}. 
While these methods have shown impressive results, they often come with significant computational costs due to their reliance on large-scale models. To address this limitation, our work focuses on accelerating these GCG models, enabling faster inference while maintaining performance.

\subsection{Token Pruning for Vision-Language Models}
The visual tokens show  more redundancy than text-tokens in visual-language models \cite{chen2024imageworth12tokensfastv}. Recently, there is a growing number of work on accelerating vision-language models (VLM) by reducing the number of visual tokens fed into the language model. One class of approaches prunes tokens based on their importance scores. In the context of llava-like VLMs, a common heuristic is to use the cross-attention between visual tokens and textual inputs: tokens that a language model attends to weakly are assumed less important and can be dropped. These works such as FastV \cite{chen2024imageworth12tokensfastv}, LLaVA-PruMerge \cite{shang2024llavaprumergeadaptivetokenreduction}
and PyramidDrop \cite{xing2025pyramiddropacceleratinglargevisionlanguage}. FastV is a notable example of this strategy. FastV~\cite{chen2024imageworth12tokensfastv} learns to prune visual tokens in the early layers of a vision-language model by identifying those tokens that have low attention impact on the text tokens. Some other token pruning methods used the input text to explicitly  guide the visual token pruning and achieve better results.\cite{zhang2025sparsevlmvisualtokensparsification, zhu2024focusllavacoarsetofineapproachefficient}. 
We diverge from prior approaches by explicitly accounting for each object’s local visual information therefore insure the object information is indeed flowed inside the visual language model.

\section{Methodology}
\begin{figure}[htbp] 
    \vspace{-6mm}
 
    \centering 
        \includegraphics[width=\linewidth]{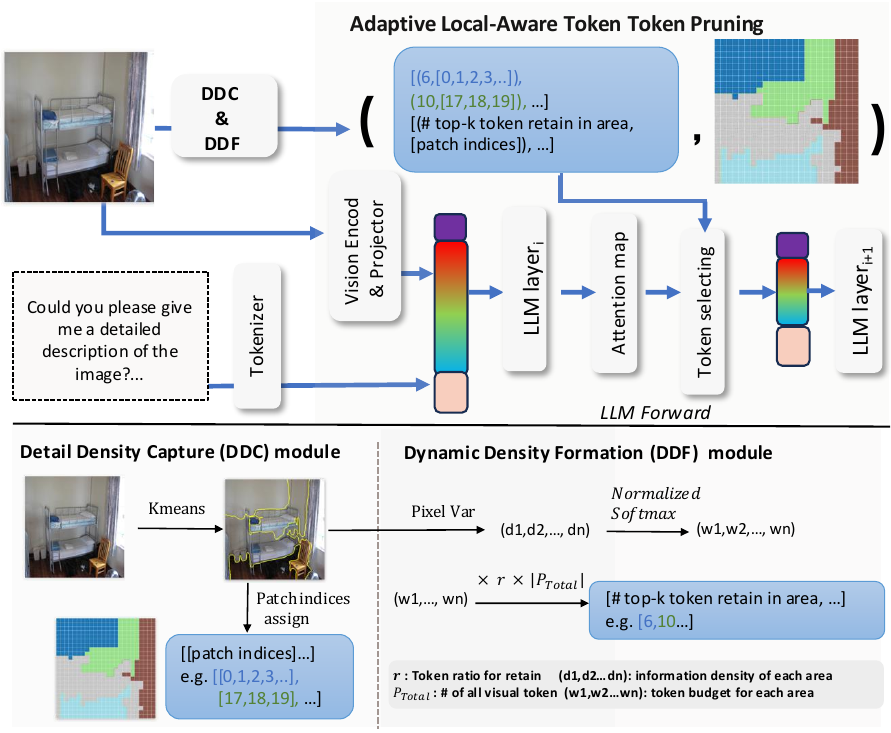} 
        \caption{Overview of the proposed Adaptive Local-Aware Token Pruning (ALTP) framework. It comprises two main components: the Detail Density Capture (DDC) module and the Dynamic Density Formation (DDF) module. The DDC module segments the image into semantically coherent sub-areas, ensuring that a larger proportion of tokens corresponding to the location of detail density regions will be retained for precise pixel-level grounding. Meanwhile, the DDF module dynamically adjusts token allocation within each region based on information density, allowing for an adaptive pruning strategy that ensuring higher token retention for tokens rich in detail. }  
        \label{fig:main_arch} 
    \vspace{-6mm}
        
\end{figure}

The  architecture overview is depicted in Figure \ref{fig:main_arch}. In the following, we first give a brief introduction of the Visual-language model and  Grounded Conversation Generation in Section \ref{sec:preli}. Then, we present our Adaptive Local-Aware Token Pruning modules in  Section \ref{sec:ddc} and Section \ref{sec:ddf}. Finally, we present the overall pruning pipeline in Section \ref{sec:over_all}. 
\subsection{Preliminaries}
\label{sec:preli}
\noindent\textbf{Visual Language Models.} Large Visual Language Models (LVLMs) are multimodal foundation models that combine visual perception capabilities with language understanding and generation. These models typically consist of three key components: (1) a visual encoder that transforms images into visual token representations, (2) a projector that aligns the visual tokens with the text embedding space, and (3) a large language model (LLM) backbone that processes the combined visual and textual information to generate responses \cite{liu2023visualinstructiontuningllava}.
Formally, given an image $I$ and a text prompt $P$, an LVLM processes the input as follows:
\begin{equation}
Res = \text{LLM}(\mathbf{Z}_v \oplus E_{\text{text}}(P)),
\\\textit{with }  \mathbf{Z}_v = Proj(g(\mathbf{X}_v)),
\end{equation}

\noindent where $g$ is the visual encoder, usually a Clip model \cite{radford2021learningtransferablevisualmodelsclip}, $\mathbf{X}_v$ is the input image, $E_{\text{text}}$ is the text tokenizer and embedding layer, $Proj$ is the vision projection layer,usually a MLP Layer, $\oplus$ denotes the concatenation operation, and $Res$ is the generated response. LVLMs excel at tasks such as image captioning, visual question answering, and multimodal reasoning, but they typically produce free-form text without explicit grounding to specific regions in the image.
\noindent\textbf{Grounded Conversation Generation}
Grounded Conversation Generation (GCG) extends the capabilities of traditional LVLMs by requiring the model to not only generate textual responses based on visual input but also to explicitly ground textual phrases to specific regions within the image.
A GCG model takes an image $I$ and a prompt $P$, which a sample prompt is like \emph{“Could you please give me a detailed description of the image? Please respond with interleaved segmentation masks for the corresponding parts of the answer.”} as input and generates a holistic   caption along with interleaved segmentation mask response $R$ along with a set of segmentation masks ${M_1, M_2, ..., M_k}$ corresponding to specific phrases ${S_1, S_2, ..., S_k}$ within the response, employing the format 
The \texttt{\textless p\textgreater} bed \texttt{\textless /p\textgreater} \texttt{\textless SEG\textgreater} is positioned beside the \texttt{\textless p\textgreater} wall \texttt{\textless /p\textgreater} \texttt{\textless SEG\textgreater}. A \texttt{\textless p\textgreater} curtain \texttt{\textless /p\textgreater} \texttt{\textless SEG\textgreater}  is
hanging from the \texttt{\textless p\textgreater} wall \texttt{\textless /p\textgreater} \texttt{\textless SEG\textgreater}. In GCG task, the start and end of each phrase and its corresponding region mask usually represent by special tokens, namely \texttt{\textless p\textgreater},\texttt{\textless /p\textgreater}   and \texttt{\textless SEG\textgreater}, respectively. The segmentation result is decoded from  last-layer embeddings corresponding to the \texttt{\textless SEG\textgreater}  tokens. In Glamm, its decoder  is a SAM-like \cite{kirillov2023segment} architecture decoder. In OMG-Llava, its decoder is comprises with cross-attention and self-attention layers, it generates the segmentation result with \texttt{\textless SEG\textgreater} tokens' embedding and image features.

\subsection{Detail Density Capture (DDC)}
\label{sec:ddc}

To retain critical local features, we propose a \textbf{Detail Density Capture} (DDC) module that segments the image into sub-regions and treats each sub-region as a unit for token pruning. Empirically, we observe that standard global attention methods tend to overlook subtle object boundaries, causing significant performance drops in GCG.

\paragraph{Superpixel-Based Sub-Regions.}
We employ superpixel segmentation (via SLIC~\cite{6205760slic}) to divide the image $\mathbf{X}_v$ into $K$ visually coherent regions:
\begin{equation}
\label{eq:slic_segment}
\{ \mathcal{S}_1, \mathcal{S}_2, \dots, \mathcal{S}_K \}
\;=\; \Phi_{\mathrm{SLIC}}\!\big(\mathbf{X}_v;N,\mathrm{C},\sigma\big),
\end{equation}
where $N$ is the target number of superpixels, $\mathrm{C}$ controls compactness (i.e., trade-off between color and spatial proximity), and $\sigma$ is a Gaussian smoothing parameter. For each $\mathcal{S}_k$, we identify the set of tokens that correspond to pixels within that superpixel. Instead of pruning solely based on global criteria, DDC ensures that each superpixel region retains at least a fraction of its tokens, thus preserving fine-grained features critical to describing local objects (e.g., ``wall,'' ``curtain,'' etc.).
\begin{figure}[htbp]
    \centering
    \includegraphics[width=\linewidth]{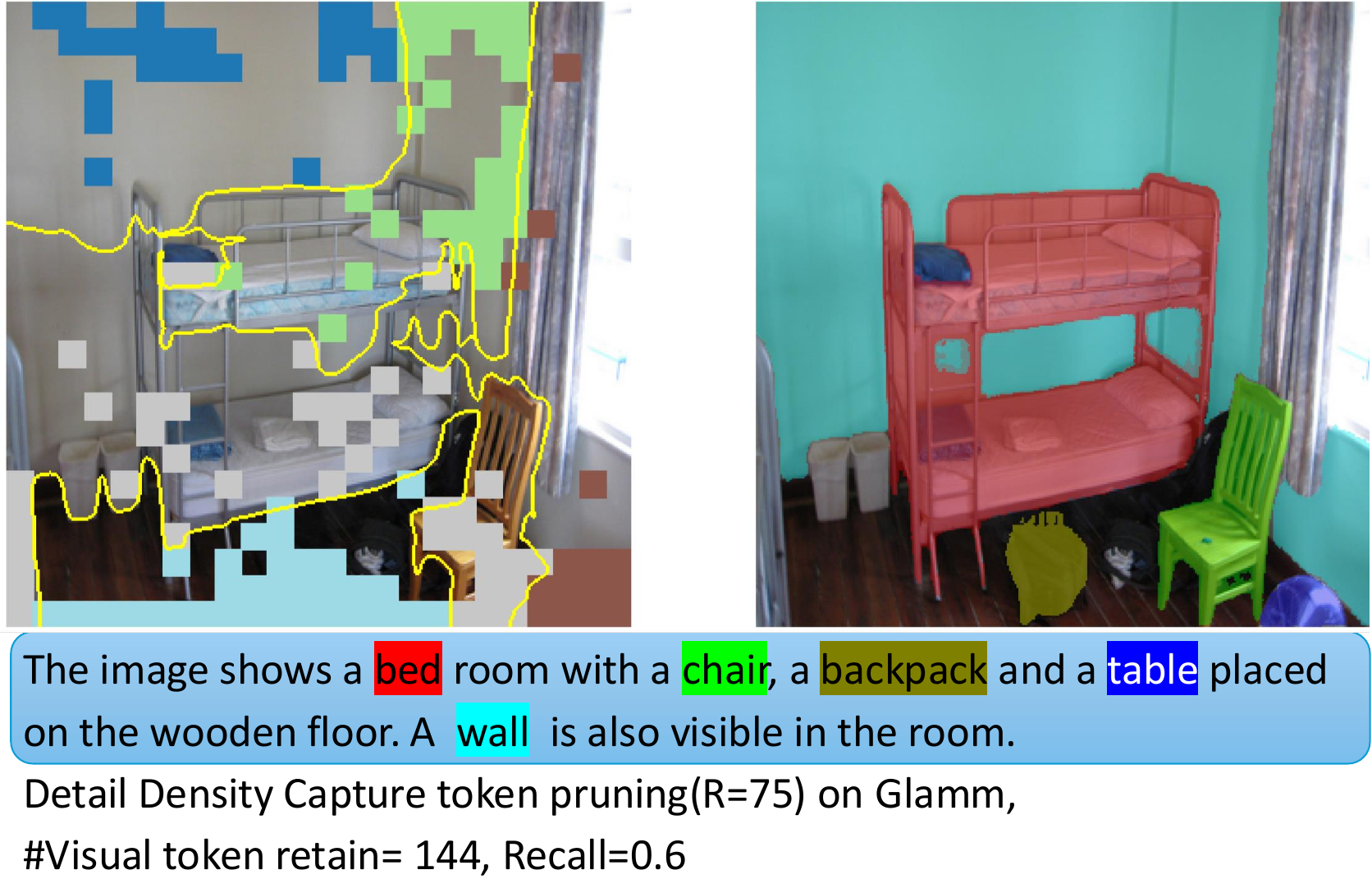}
    \vspace{-6mm}
    \caption{Detail Density Capture (DDC) visualization. Left: Retained token locations using DDC with a 75\% token drop. Right: Grounded conversation generation result using DDC, demonstrating successful generation of the "wall" phrases and mask.}
    \label{fig:ddc_vis}
    \vspace{-6mm}
\end{figure}
\paragraph{Pruning Within Each Sub-Region.}
Let $\Omega_k$ be the set of visual-token indices corresponding to superpixel $\mathcal{S}_k$. We maintain a local keep-ratio $r_k$ for each region. If $|\Omega_k|$ is the number of tokens in region $k$, we keep $\lceil r_k \cdot |\Omega_k| \rceil$ tokens. The simplest version of DDC sets $r_k$ identically for all $k$, ensuring uniform retention across sub-regions. However, uniform distribution still may not reflect varying complexity among objects (e.g., large uniform backgrounds vs.\ small, detail-rich objects). This motivates our second module, DDF, to allocate tokens \emph{dynamically}.

\subsection{Dynamic Density Formation (DDF)}
\label{sec:ddf}

Although DDC preserves local features, it treats each region equally. In practice, large objects (e.g., walls) may not require as many tokens as smaller, high-detail objects (e.g., a lamp with intricate shapes). Hence, we introduce a \textbf{Dynamic Density Formation (DDF)} strategy to allocate the overall token budget proportionally to a region's \emph{information density}.

\paragraph{Information Density.}
For each superpixel $\mathcal{S}_k$, we define an \emph{information density} $d_k$ that combines pixel variance and area size:
\begin{equation}
\label{eq:ddf_density}
d_k \;=\; \mathrm{Var}\big(\mathcal{S}_k\big)\;\sqrt{\frac{|\mathcal{P}_k|}{|\mathcal{P}_\mathrm{total}|}},
\end{equation}
where $\mathrm{Var}(\mathcal{S}_k)$ is the color variance within region $k$, $|\mathcal{P}_k|$ is the number of pixels (or patches) in $\mathcal{S}_k$, and $|\mathcal{P}_\mathrm{total}|$ is the total number of pixels in the entire image. The multiplier $\sqrt{\frac{|\mathcal{P}_k|}{|\mathcal{P}_\mathrm{total}|}}$ prevents extremely small but high-variance regions from dominating the token budget.

\begin{figure}[htbp]
    \centering
    \includegraphics[width=\linewidth]{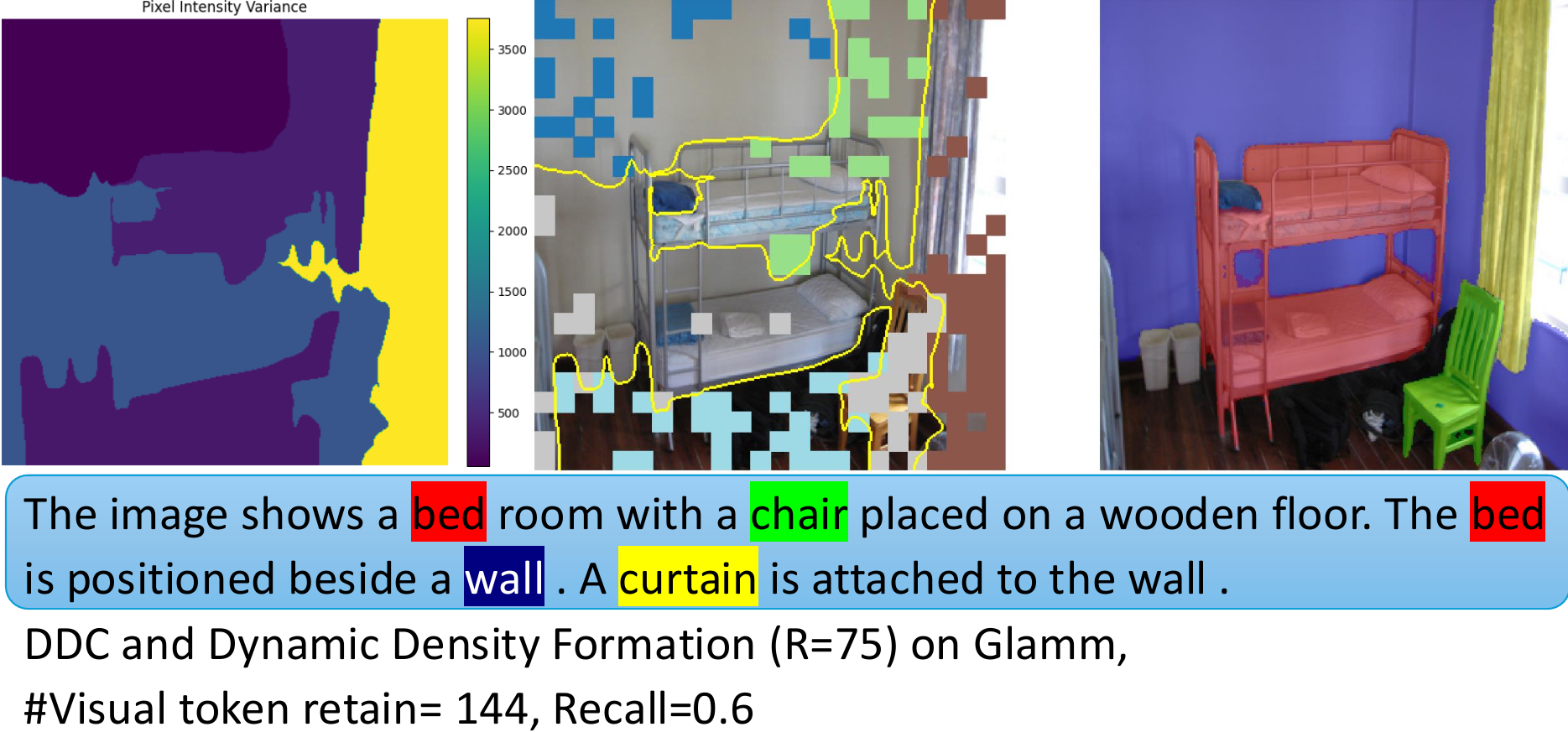}
    \caption{Visualization of Dynamic Density Formation (DDF) token allocation. \textbf{Left}: Pixel variance for each sub-area calculated using Equation \ref{eq:ddf_density}, indicating higher information density in regions like the curtain. \textbf{Middle}: Corresponding token allocation weights derived from information density via Equation \ref{eq:ddf_weight}, showing that tokens in regions with higher variance (e.g., the curtain) receive a larger allocation budget. \textbf{Right}: Final generation results with the DDC and DDF modules under a 25\% token allocation setting. Compared to uniform token allocation in DDC as shown in Figure \ref{fig:ddc_vis}, the DDF module enables the model to dynamic allocate token budge to the tokens with more information density, therefor, capture and generate detailed objects (e.g., the curtain). } 
    \label{fig:ddf_vis}
    \vspace{-6mm}
\end{figure}

\paragraph{Token Allocation.}
Given a total keep-ratio $r$ for visual tokens, we wish to distribute these tokens among the $K$ superpixels proportionally to $d_k$. We first compute normalized weights $w_k$ via a softmax-like mechanism:
\begin{equation}
\label{eq:ddf_weight}
w_k \;=\; \frac{\exp\Big(\tfrac{d_k}{\,\alpha \cdot \max(d)\,}\Big)}{\sum_{j=1}^{K}\exp\Big(\tfrac{d_j}{\,\alpha \cdot \max(d)\,}\Big)},
\end{equation}
where $\alpha > 1$ is a scaling factor (e.g., $\alpha = 1.5$) to avoid overly skewed allocations. Then, if $V_\mathrm{total}$ is the initial number of visual tokens, we allocate:
\[
T_k \;=\; w_k \times \bigl(r \,\cdot\, V_\mathrm{total}\bigr)
\]
tokens to superpixel $\mathcal{S}_k$. Concretely, we keep the top $\lceil T_k \rceil$ tokens (by some local importance criterion, e.g.\ cross-attention or self-attention magnitude) within that superpixel. Figure~\ref{fig:ddf_vis} shows that DDF adaptively allocates more tokens to high-detail objects (e.g., a curtain), improving segmentation accuracy compared to uniform allocation.

\subsection{Overall Pruning Pipeline}
\label{sec:over_all}

Algorithm~\ref{alg:main} summarizes our full approach. We first generate superpixels via DDC (Eq.~\ref{eq:slic_segment}), then compute $d_k$ for each sub-region (Eq.~\ref{eq:ddf_density}). We convert $d_k$ into allocation weights $w_k$ (Eq.~\ref{eq:ddf_weight}) and select tokens accordingly. This pruned set of tokens is concatenated with the text embedding and fed into the language model.

\begin{algorithm}[t]
\caption{Adaptive Local-Aware Token Pruning (ALTP)}
\label{alg:main}
\begin{algorithmic}[1]
\Require Image $\mathbf{X}_v$, Prompt $\mathbf{P}$, Total Keep-Ratio $r$, Number of Superpixels $N$, Compactness $C$
\State \emph{// \textbf{Step 1: DDC}}
\State $\{\mathcal{S}_1,\ldots,\mathcal{S}_K\} \leftarrow \Phi_{\mathrm{SLIC}}(\mathbf{X}_v;N,C,\sigma)$
\For{$k=1$ to $K$}
    \State Identify token indices $\Omega_k$ of $\mathcal{S}_k$
\EndFor
\State \emph{// \textbf{Step 2: DDF}}
\For{$k=1$ to $K$}
    \State Compute $d_k = \mathrm{Var}(\mathcal{S}_k)\,\sqrt{\frac{|\mathcal{P}_k|}{|\mathcal{P}_\mathrm{total}|}}$
\EndFor
\State Compute $w_k$ using Eq.~\eqref{eq:ddf_weight} for $k=1,\dots,K$
\State \emph{// \textbf{Step 3: Token Selection}}
\State $V_\mathrm{total} \leftarrow$ total \# of visual tokens from Enc$_v$
\For{$k=1$ to $K$}
    \State $T_k \leftarrow w_k\,\times\,(r \cdot V_\mathrm{total})$
    \State Keep top $\lceil T_k \rceil$ tokens in $\Omega_k$ based on local importance
\EndFor
\State $\Omega_\mathrm{selected} \leftarrow \bigcup_{k=1}^{K}$ (kept tokens in $\Omega_k$)
\State \emph{// \textbf{Step 4: Inference}}
\State Forward $\mathrm{E_{text}}(\mathbf{P}) \oplus \mathrm{Proj}(\mathrm{Enc_v}(\Omega_\mathrm{selected}))$ to LLM
\end{algorithmic}
\end{algorithm}

By combining DDC and DDF, our ALTP framework ensures that tokens corresponding to high-detail or semantically rich objects are preserved, while uniform or low-detail areas are pruned aggressively.

\section{Experiment results}
\subsection{Experimental Setup}

\noindent \textbf{Datasets and Evaluation Metrics.} We evaluate our proposed token pruning method on two state-of-the-art Grounded Conversation Generation (GCG) models: Glamm \cite{rasheed2024glammpixelgroundinglarge} and OMG-Llava \cite{zhang2024omgllavabridgingimagelevelobjectlevel}. Experiments are conducted using the GranDf dataset, the only established benchmark for GCG, with 2.5K validation and 5K test samples. Performance is assessed using : METEOR and CIDEr for captioning quality, AP50, mask Intersection over Union (IoU) for segmentation mask, and mask recall for evaluating region-specific grounding. The recall calculation follows that of GlaMM, counting a “hit” only if the generated mask and phrase pair surpass both 0.5 IoU and 0.5 BERT similarity thresholds with respect to the ground-truth mask and phrases. 

\noindent \textbf{FLOPS estimated} 
 For one transformer layer with multi-head attention and feed-forward network (FFN), assume $n$ is the token number, $d$ is the hidden state size, $m$ is the intermediate size of FFN, the total FLOPs can be estimated by $f_l(n)=4nd^2+2n^2d+2ndm$. For the whole model, assume  pruning tokens from $n$ to $\hat{n}=(1-R\%)\cdot n$ after layer $K$ and there are T layers  all. The theoretical FLOPs reduction ratio related to image tokens is computed as:
\begin{equation}
    1-\frac{K\times f_l(n)+(T-K)\times f_l(\hat{n})}{T\times f_l(n)}
\end{equation}
% For one transformer layer of Group Query Attention (GQA), which is used in OMG-Llava, with  $H$ total heads and $G$ groups, the FLOPs for one layer can be estimated as:
% $f_{GQA}(n) = (2+\frac{2G}{H})nd^2 + 2n^2d + 2ndm$.
% Here, $(2+\frac{2G}{H})nd^2$ accounts for Query projections and shared Key/Value projections across $G$ groups. The terms $2n^2d$ and $2ndm$ represent the scaled dot-product attention and FFN module FLOPs, respectively, similar to MHA.

\begin{table*}[]
\renewcommand{\arraystretch}{0.9}
\setlength{\tabcolsep}{5pt} % 调整列间距

\begin{tabular}{llllllllllll}
\toprule
\multicolumn{1}{c|}{}                               & \multicolumn{1}{c|}{}                        & \multicolumn{5}{c|}{Val}                                                                                                                   & \multicolumn{5}{c}{Test}                                                                                                                                                       \\ \cline{3-12} 
\multicolumn{1}{c|}{}                               & \multicolumn{1}{c|}{\multirow{-2}{*}{FLOPS}} & METEOR        & CIDEr         & AP50                                 & mIOU                        & Recall                               & METEOR                      & CIDEr                       & AP50                                 & mIOU                                 & Recall                               \\ \cline{2-12} 
\multicolumn{1}{c|}{}                               & \multicolumn{11}{c}{Upper Bound, Retain   576 Tokens (100\%)}                                                                                                                                                                                                                                                                                                            \\ \cline{2-12} 
\multicolumn{1}{c|}{\multirow{-4}{*}{Model   name}} & 100\%                                       & 16.2          & 47.2          & 30.8                                 & 66.3                        & 41.8                                 & 15.8                        & 43.5                        & 29.2                                 & 65.6                                 & 40.8                                 \\ \hline
\multicolumn{12}{c}{Retain  57 Tokens ( \textcolor{darkgreen}{$\downarrow$ 90\%}
 )}  \\ \hline
Fastv                                               & 20\%                                        & 12.6          & 27.8          & 14.0                                 & 48.1                        & 23.3                                 & 13.1                        & 28.3                        & 13.6                                 & 47.9                                 & 25.1                                 \\
PDrop                                               & 20\%                                        & \textbf{12.8} & \textbf{28.1} & 15.1                                 & 49.7                        & 25.3                                 & \textbf{12.9}               & 28.7                        & 14.6                                 & 59.1                                 & 26.8                                 \\
ALTP                                                & 20\%                                        & 12.4          & \textbf{28.1} & \textbf{20.0}                        & \textbf{55.5}               & \textbf{30.3}                        & 12.0                        & \textbf{29.0}               & \textbf{19.4}                        & \textbf{54.6}                        & \textbf{34.4}                        \\ \hline
\multicolumn{12}{c}{Retain  144 Tokens ( \textcolor{darkgreen}{$\downarrow$ 75\%}
)}      \\ \hline
Fastv                                               & 33\%                                        & 12.8          & 30.9          & 20.9                                 & 50.8                        & 30.4                                 & 12.7                        & 29.0                          & 20.2                                 & 53.8                                 & 30.0                                   \\
PDrop                                               & 33\%                                        & \textbf{13.5} & \textbf{31.5} & 21.7                                 & 52.1                        & 31.9                                 & \textbf{13.1}               & \textbf{30.5}               & 21.5                                 & 55.2                                 & 31.7                                 \\
ALTP                                                & 33\%                                        & 13.4          & 31.1          & \textbf{24.8}                        & \textbf{60.3}               & \textbf{34.2}                        & 12.4                        & 30.1                        & \textbf{23.9}                        & \textbf{59.7}                        & \textbf{33.1}                        \\ \hline
\multicolumn{12}{c}{Retain  288 Tokens (\textcolor{darkgreen}{$\downarrow$ 50\%}
)}    \\ \hline
Fastv                                               & 55\%                                        & 13.2          & 34.3          & 24.4                                 & 61.9                        & 33.2                                 & 13.1                        & 32.1                        & 23.3                                 & 61.2                                 & 32.9                                 \\
PDrop                                               & 55\%                                        & \textbf{13.8} & \textbf{34.1} & 25.6                                 & \textbf{62.7}               & 33.9                                 & 14.2                        & \textbf{33.7}               & 24.1                                 & 61.9                                 & 33.0                                 \\
ALTP                                                & 55\%                                        & 13.6          & 33.7          & {\color[HTML]{333333} \textbf{26.3}} & {\color[HTML]{333333} 62.2} & {\color[HTML]{333333} \textbf{34.9}} & {\color[HTML]{333333} 13.4} & {\color[HTML]{333333} 33.4} & {\color[HTML]{333333} \textbf{25.1}} & {\color[HTML]{333333} \textbf{62.1}} & {\color[HTML]{333333} \textbf{34.4}} \\ \bottomrule
\end{tabular}
\vspace{-3mm}
\caption{Result of our pruning method on GlaMM on GranDf dataset under different setting of retain/pruning token ratio.  ALTP significantly outperforms Pdrop \cite{xing2025pyramiddropacceleratinglargevisionlanguage} and Fastv\cite{chen2024imageworth12tokensfastv} under  75\% and 90\% token dropping setting. }
\label{tab:glamm-result}
\end{table*}
\noindent \textbf{Comparison Methods.} We benchmark our proposed token pruning method against existing state-of-the-art methods, including FastV \cite{chen2024imageworth12tokensfastv} and PDrop \cite{xing2025pyramiddropacceleratinglargevisionlanguage}. The choice is   driven by two three factors.   First, other widely-used datasets such as refCOCO, refCOCO+, and refCOCOg \cite{Kazemzadeh2014ReferItGameRTrefcocog} are primarily designed for text-driven segmentation tasks, where the model outputs a segmentation mask based solely on a given query text, disregarding phrase generation. Moreover, state-of-the-art methods on these datasets, such as LISA \cite{lai2024lisa}, generate segmentation masks directly through decoding special tokens (e.g.,  \texttt{\textless SEG\textgreater}  tokens) in a non-autoregressive manner, fundamentally differing from Grounded Conversation Generation (GCG) models like Glamm, which produce both phrases and segmentation masks auto-regressively. Consequently, such datasets and models are not directly comparable with GCG models. Second, our proposed ALTP method specifically targets visual token pruning in large visual-language models architecturally similar to Llava. Lastly, some other token pruning methods used the relevant text in query  to explicitly  guide the visual token pruning and achieve better results.\cite{zhang2025sparsevlmvisualtokensparsification, zhu2024focusllavacoarsetofineapproachefficient}. These methods do not fit the GCG task because the input text of GranDf dataset does not have relevant text information that could benefit token pruning. A sample query is  “Could you please give me a detailed description of the image? Please respond with interleaved segmentation masks for the corresponding parts of the answer.” Thus, our experimental comparison appropriately focuses on methods aligned with this scenario.

\subsection{Experiments on GCG task}
\noindent \textbf{ALTP excels with  Glamm on GranDf dataset} 
As Table~\ref{tab:glamm-result} presents the performance of our proposed ALTP method alongside FastV \cite{chen2024imageworth12tokensfastv} and PDrop \cite{xing2025pyramiddropacceleratinglargevisionlanguage} on the Glamm model, evaluated on the GranDf dataset. We assess the models across various token retention ratios, specifically retaining 57, 144, and 288 tokens, corresponding to a 90\%, 75\%, and 50\% reduction in visual tokens, respectively. 
At the highest compression ratio, retaining only 57 tokens (90\% reduction), ALTP achieves remarkable gains in structural understanding metrics. Specifically, ALTP improveS AP50 by \textbf{4.9\%} (20.0 vs. 15.1) and Recall by \textbf{5.0\%} (30.3 vs. 25.3) on the Val set compared to PDrop. This performance advantage is even more pronounced when compared to FastV. When retaining 144 tokens (75\% reduction), ALTP attain the highest mIOU of 60.3\% on Val, outperforming PDrop by \textbf{8.2\%} (60.3 vs. 52.1) and FastV by \textbf{9.5\%} (60.3 vs. 50.8).  With 288 tokens retained (50\% reduction), the performance gap between ALTP and PDrop narrows. Specifically, ALTP achieves 26.3\% AP50 and 62.2\% mIOU on the validation set, and 25.1\% AP50 and 62.1\% mIOU on the test set. 
In summary, though PDrop shows slightly strong performance in captioning metrics,  ALTP demonstrates superior performance in visual metrics  across varying token retention ratios, particularly at high compression rates.

\noindent \textbf{ALTP perform well with  OMG-Llava } Table~\ref{tab:result_omg_llava} demonstrates the efficacy of ALTP on  the OMG-Llava model. With 25 tokens retained (90\% reduction), ALTP significantly enhances performance. Specifically, our method improves AP by \textbf{2.1\%} and mIOU by \textbf{3.0\%} on the validation set compared to PDrop. On the test set,ALTP improves AP by 1.5\% and mIOU by 1.6\%. This highlights the superior grounding accuracy and segmentation precision achieved by our approach under high compression. At 64 tokens (75\% reduction), ALTP maintains a competitive edge, improving AP by 0.7\% on the validation set and 0.6\% on the test set, while demonstrating comparable mIOU and Recall scores. This confirms our method's ability to preserve crucial visual information for grounded conversation generation, especially under severe token reduction. The potential reason that our method’s 
performance improvement is more pronounced when applied to Glamm than to OMG-Llava is discussed in Section \ref{sec:discussion} .

\begin{table}[]
\setlength{\tabcolsep}{3pt} % 调整列间距
\renewcommand{\arraystretch}{1.3} % 调整行间距
\renewcommand{\arraystretch}{1}
\centering
\begin{tabular}{c|ccc|ccc}
\hline
Model Name & \multicolumn{3}{c|}{Val} & \multicolumn{3}{c}{Test} \\ \hline
           & AP    & mIOU  & R    & AP    & mIOU  & R \tablefootnote{Note that the original OMG-Llava paper does not calculate recall metrics. The recall in this table is result from our experiment.}    \\ \hline
\multicolumn{7}{c}{Upper Bound, Retain 576 Tokens (100\%)} \\ \hline
           & 29.9  & 65.5  & 43.6 & 29.2  & 64.6  & 42.3 \\ \hline
\multicolumn{7}{c}{Retain 25 Tokens ( \textcolor{darkgreen}{$\downarrow$ 90\% \textcolor{black}{)}}} \\ \hline
Pdrop      & 24.1  & 58.0  & 35.4 & 22.9  & 57.7  & 34.1 \\
ALTP      & \textbf{26.2} & \textbf{61.0}   & \textbf{37.4} & \textbf{24.4} & \textbf{59.3} & \textbf{35.6} \\ \hline
\multicolumn{7}{c}{Retain 64 Tokens ( \textcolor{darkgreen}{$\downarrow$ 75\% \textcolor{black}{)}}} \\ \hline
Pdrop      & 27.2  & 63.5  & 38.7 & 26.3  & 63.2  & 38.6 \\
ALTP                                               & \textbf{27.9} & \textbf{63.6} & \textbf{39.2} & \textbf{26.9} & \textbf{62.3} & \textbf{39.5} \\ \hline
\end{tabular}
\caption{Result of our pruning method on OMG-Llava on GranDf dataset under different setting of retain/pruning token ratio. AP, R represent for AP50 and recall respectively. ALTP outperforms Pdrop under 75\% and 90\% token dropping.}
\vspace{-5mm}
\label{tab:result_omg_llava}
\end{table}

\subsection{Module Ablation study}
Table~\ref{tab:abla_module} presents an ablation study evaluating the impact of our ALTP method's components, Detail Density Capture (DDC) and Dynamic Density Formation (DDF). At a 90\% token reduction (57 tokens), ALTP achieves a 20.0\% AP and 55.5\% mIOU on the validation set, compared to 18.9\% AP and 53.8\% mIOU with only DDC. This indicates that DDF contributes a 1.1\% improvement in AP and a 1.7\% increase in mIOU. On the test set, ALTP improves Recall by 5.0\% over only using DDC. With 144 tokens retained (75\% reduction), ALTP shows a 0.5\% AP and 0.2\% mIOU improvement on the validation set, demonstrating that DDF fine-tunes the token distribution for better performance. At 288 tokens (50\% reduction), ALTP achieves a 26.3\% AP, compared to 25.5\% with only DDC, a 0.8\% improvement. These results validate that while DDC provides a strong baseline by focusing on local information, DDF further enhances performance by dynamically allocating tokens based on information density.

\begin{table}[]
\setlength{\tabcolsep}{3pt} % 调整列间距
\renewcommand{\arraystretch}{1.3} % 调整行间距
\renewcommand{\arraystretch}{1}
\begin{tabular}{ccccccccccccc}
\cline{1-7}
\multicolumn{1}{c|}{\multirow{2}{*}{Module name}} & \multicolumn{3}{c|}{Val}                               & \multicolumn{3}{c}{Test}                      &  &  &  &  &  &  \\ \cline{2-7}
\multicolumn{1}{c|}{}                        & AP            & mIOU          & \multicolumn{1}{c|}{R} & AP            & mIOU          & R             &  &  &  &  &  &  \\ \cline{1-7}
\multicolumn{7}{c}{Retain  57 Tokens (\%)}                                                                                                            &  &  &  &  &  &  \\ \cline{1-7}
Only DDC                                     & 18.9          & 53.8          & 29.1                   & 18.8          & 54.2          & 29.4          &  &  &  &  &  &  \\
ALTP                                         & \textbf{20.0}   & \textbf{55.5} & \textbf{30.3}          & \textbf{19.4} & \textbf{54.6} & \textbf{34.4} &  &  &  &  &  &  \\ \cline{1-7}
\multicolumn{7}{c}{Retain  144 Tokens (75\%)}                                                                                                         &  &  &  &  &  &  \\ \cline{1-7}
Only DDC                                     & 24.3          & 60.1          & 33.8                   & 23.7          & 59.2          & \textbf{33.1} &  &  &  &  &  &  \\
ALTP                                         & \textbf{24.8} & \textbf{60.3} & \textbf{34.2}          & \textbf{23.9} & \textbf{59.7} & \textbf{33.1} &  &  &  &  &  &  \\ \cline{1-7}
\multicolumn{7}{c}{Retain  288 Tokens (50\%)}                                                                                                         &  &  &  &  &  &  \\ \cline{1-7}
Only DDC                                     & 25.5          & 62.1          & 34.8                   & 24.7          & \textbf{62.1} & 34.5          &  &  &  &  &  &  \\
ALTP                                         & \textbf{26.3} & \textbf{62.2} & \textbf{34.9}          & \textbf{25.1} & \textbf{62.1} & \textbf{34.6} &  &  &  &  &  &  \\ \cline{1-7}
\end{tabular}
\caption{Ablation study showing that while DDC effectively preserves local information, DDF's dynamic allocation based on information density further improves grounding and segmentation metrics.}
\label{tab:abla_module}
\vspace{-5mm}
\end{table}

\subsection{Hyperparameter robustness study}
Table~\ref{tab:hyper_robust} examines the robustness of our ALTP method to variations in superpixel hyperparameters, specifically the  number of areas (N) and compactness (C). When N is set to 3, a significant performance drop is observed, with AP decreasing to 17.1\% on both validation and test sets. This occurs because with very few sub-areas, ALTP's localized pruning reverts to a more global, attention-based approach akin to FastV. Optimal performance is achieved with N=7, yielding the highest AP (20.5\% on validation, 19.8\% on test) and mIOU (55.9\% on validation, 54.8\% on test). For C, variations between 3, 5 (default), and 10 show minimal impact, indicating ALTP's stability across a range of compactness values. This suggests that while the number of sub-areas  affects performance, the balance between color and spatial proximity within superpixels is less critical.

\begin{table}[]
\setlength{\tabcolsep}{3pt} % 调整列间距
\renewcommand{\arraystretch}{1.3} % 调整行间距
\renewcommand{\arraystretch}{1}
\begin{tabular}{ccccccccccccc}
\hline
\multicolumn{2}{c|}{\multirow{2}{*}{\begin{tabular}[c]{@{}c@{}}Hyper-\\ parameter\end{tabular}}} & \multicolumn{3}{c|}{Val}                               & \multicolumn{3}{c}{Test}                      \\
\multicolumn{2}{c|}{}                                                                            & AP            & mIOU          & \multicolumn{1}{c|}{R} & AP            & mIOU          & R             \\ \hline
\multirow{4}{*}{N}                                 & 3                                           & 17.1          & 53.7          & 27.1                   & 17.1          & 51.8          & 27.6          \\
                                                   & 7                                           & \textbf{20.5} & \textbf{55.9} & \textbf{30.8}          & \textbf{19.8} & \textbf{54.8} & \textbf{34.7} \\
                                                   & 10 (Default)                                & 20            & 55.5          & 30.3                   & 19.4          & 54.6          & 34.4          \\
                                                   & 15                                          & 20.1          & 55.4          & 30.6                   & 19.1          & 54.7          & 34.3          \\ \hline
\multirow{3}{*}{C}                                 & 3                                           & 20.2          & 55.3          & \textbf{30.5}          & 19.1          & 54.5          & 34.2          \\
                                                   & 5 (Default)                                 & 20            & 55.5          & 30.3                   & 19.4          & \textbf{54.6} & \textbf{34.4} \\
                                                   & 10                                          & 19.8          & \textbf{55.7} & 30.1                   & \textbf{19.6} & 54.4          & 34.3          \\ \hline
\end{tabular}
\caption{Analysis of Hyperparameter   N (Target Number of Areas) and C (Compactness) on ALTP Performance. The results show ALTP is robust to N from 7 to 15 and C from  3 to 10. }
\label{tab:hyper_robust}
\end{table}

\section{Discussion}
\label{sec:discussion}
\begin{figure}[htbp] % 使用 figure 环境来包含图片
    \centering % 图片居中
        \includegraphics[width=\linewidth]{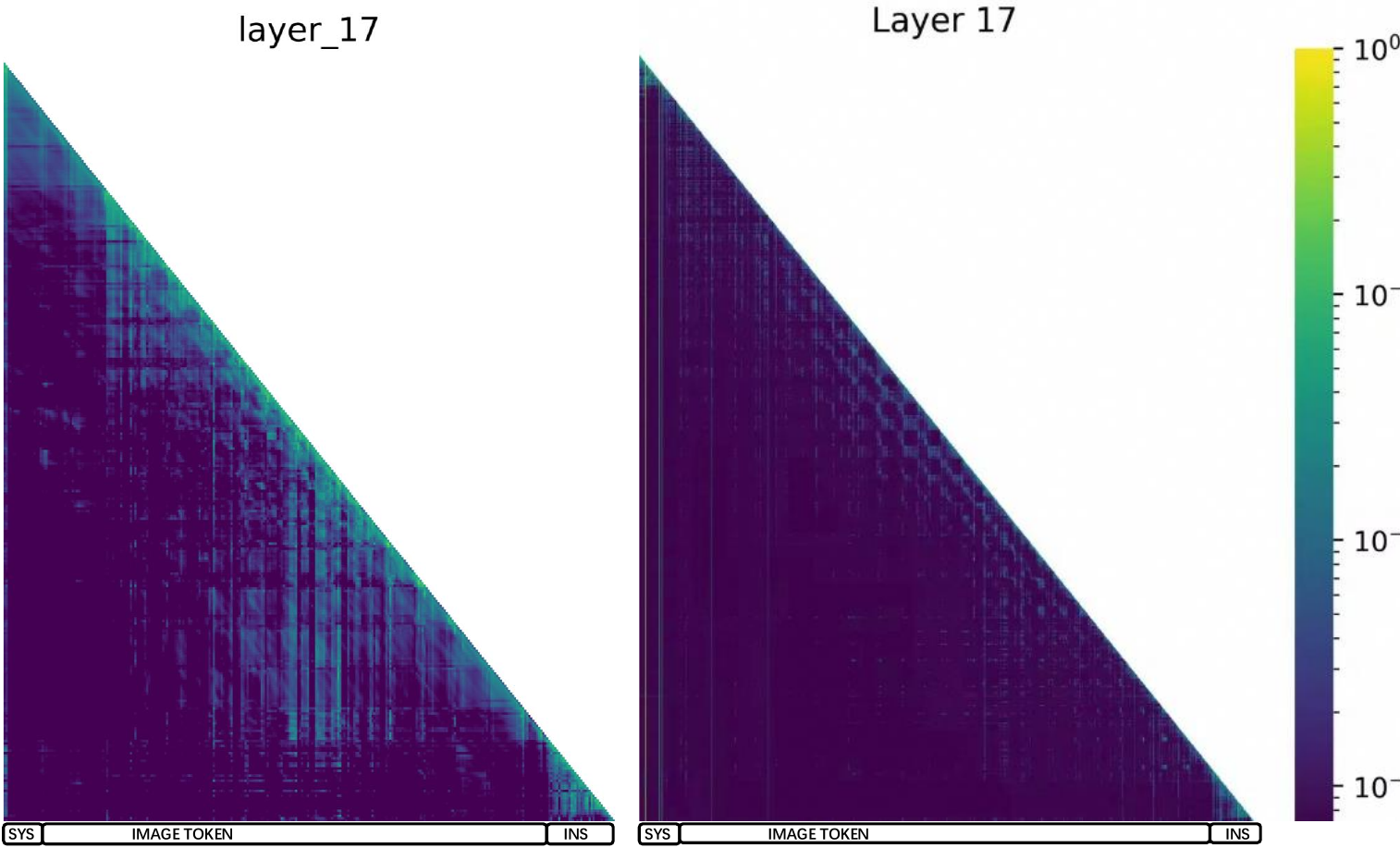} % 包含第一个 PDF 图片
        \caption{Left: The attention map of OMG-Llava on Grand 
 \cite{rasheed2024glammpixelgroundinglarge} dataset. Right: The attention map of Llava. We could observe a huge difference among the attention patterns of  image tokens. }  
        \label{fig:ATTENTION} 
    \vspace{-4mm}
        
\end{figure}

Figure \ref{fig:ATTENTION} illustrates the attention maps of the middle layers during the decoding process for OMG-Llava and Glamm. Visual analysis reveals a significant disparity in image token attention between the two models when evaluated on the Grandf [23] dataset. Specifically, OMG-Llava's attention is more concentrated on image tokens corresponding to the image's center, compared to Glamm. Given that center image regions typically contain more salient objects, pruning methods like Fastv or PDrop  naturally retain these central tokens. This tendency partially explains why our method's performance improvement is more pronounced when applied to Glamm than to OMG-Llava.

\section{Conclusion}
We present Adaptive Local-Aware Token Pruning (ALTP), a novel framework that addresses the computational burden of Grounded Conversation Generation (GCG) models by prioritizing local visual features during token pruning. Our ALTP integrates Detail Density Capture (DDC), which segments images and retains tokens based on regional importance, and Dynamic Density Formation (DDF), which dynamically allocates tokens according to information density. Through extensive experiments on GLaMM and OMG-LLaVA, we demonstrate that ALTP significantly outperforms existing token pruning methods. ALTP provides a promising approach for accelerating GCG models by effectively managing local information crucial for accurate visual grounding.
\clearpage

{
    \small
    \bibliographystyle{ieeenat_fullname}
    \bibliography{main}
}

\end{document}